\documentclass[journal,twocolumn]{IEEEtran}
\usepackage{xcolor}
\usepackage{mathtools}
\usepackage{epsfig,makeidx,color,subcaption}
\usepackage{amsmath,amssymb,bbm,enumitem}
\usepackage{cite,graphicx,lipsum}
\usepackage{booktabs}
\usepackage{multirow}
\usepackage{tablefootnote}
\usepackage[switch,pagewise]{lineno}
\usepackage{hyperref}
\hypersetup{
        colorlinks = true,
        citecolor=red,
}

\def\cE{{\cal E}}
\def\cM{{\cal M}}

\def\cK{{\cal K}}

\def\rT{{\rm T}}

\def\uR{{\mathbb R}}

\def\be{ \begin{equation} }
\def\ee{ \end{equation} }
\def\bea{ \begin{eqnarray} }
\def\eea{ \end{eqnarray} }

\def\bx{{\bf x}}

\def\bb{{\bf b}}

\def\bg{{\bf g}}

\def\bu{{\bf u}}

\def\bz{{\bf z}}

\def\bff{{\bf f}}
\def\bv{{\bf v}}

\def\bF{{\bf F}}
\def\bG{{\bf G}}

\def\bI{{\bf I}}

\def\bK{{\bf K}}

\def\bV{{\bf V}}

\def\bU{{\bf U}}

\def\b0{{\bf 0}}

\def\bSigma{{\bf \Sigma}}

\def\cL{{\cal L}}
\def\cD{{\cal D}}

\begin{document}

\title{Koopman AutoEncoder via Singular Value Decomposition for Data-Driven Long-Term Prediction}

\author{Jinho Choi, Sivaram Krishnan, and Jihong Park
\thanks{
J. Choi and S. Krishnan are with Deakin University and J. Park is with
 Singapore University of Design and Technology.
This research was supported by the Defence Science and Technology Group (DSTG), Australia. We would like to thank Gregory Sherman and Benjamin Campbell from DSTG for their valuable comments and suggestions that greatly improved the quality of this paper.}}

\maketitle

\begin{abstract}
The Koopman autoencoder, a data-driven technique, has gained traction for modeling nonlinear dynamics using deep learning methods in recent years. Given the linear characteristics inherent to the Koopman operator, controlling its eigenvalues offers an opportunity to enhance long-term prediction performance, a critical task for forecasting future trends in time-series datasets with long-term behaviors. However, controlling eigenvalues is challenging due to high computational complexity and difficulties in managing them during the training process.
To tackle this issue, we propose leveraging the singular value decomposition (SVD) of the Koopman matrix to adjust the singular values for better long-term prediction. Experimental results demonstrate that, during training, the loss term for singular values effectively brings the eigenvalues close to the unit circle, and the proposed approach outperforms existing baseline methods for long-term prediction tasks.
\end{abstract}

\section{Introduction}
\label{sec:intro}

Koopman operator theory posits that smooth nonlinear dynamics can be represented as globally linear dynamics within a high-dimensional (or infinite-dimensional) space \cite{koopman1931hamiltonian, Brunton_PLOS}. This approach has been actively studied for modeling nonlinear dynamics \cite{Brunton22}. Additionally, deep learning techniques, such as the encoder-decoder structured neural network (NN) known as the Koopman autoencoder (KAE), can be employed \cite{lusch2018deep} \cite{Takeishi17}. The KAE also holds great potential for long-term prediction, in contrast to conventional nonlinear prediction methods that rely on locally linear approximations at specific reference points and are accurate only for short-term prediction.

The KAE consists of three key blocks: the encoder, the decoder, and the \emph{Koopman matrix}. While the encoder and decoder are NNs, the Koopman matrix is a square matrix that functions as a linear layer inserted between the two NNs. The Koopman matrix plays a crucial role in determining dynamics in a linear subspace, and $\tau$-step prediction boils down to repeatedly multiplying its eigenvalues $\tau$ times. Hence, it is necessary to keep the eigenvalues neither too small nor too large for long-term prediction. To achieve this, the consistent KAE (CKAE) in \cite{Azencot20a} enforces consistency between forward and backward predictions, which \emph{implicitly} controls the eigenvalues within a favorable range.

Alternatively, in this paper, by leveraging the relationship between eigenvalues and singular value decomposition (SVD), we propose a novel KAE architecture termed unrolled SVD-based CKAE (USVD-CKAE). In USVD-CKAE, the Koopman matrix is inherently decomposed into three SVD components, which effectively control the eigenvalues during training while avoiding eigendecomposition or SVD computations in each training round. Numerical simulations with fluid flow dynamics demonstrate that USVD-CKAE achieves up to 66.43\% and 91.01\% lower 1000-step future prediction errors compared to Vanilla KAE and CKAE, respectively.

\section{Backgrounds and Related Works}

\subsection{Koopman Operator Theory}
Consider the following discrete-time dynamical system:
\be 
\bx(t+1) = \bF (\bx(t)) \in \cM \subseteq \uR^N,
    \label{EQ2:xFx}
\ee 
where $\bx(t)$ is the state vector of length $N$, $\cM$ is the state space, and $\bF: \uR^N \to \uR^N$ is the flow map.
For a given map~$\bF$, there exists a \emph{linear} operator $\cK$ that acts to advance all observables $g: \cM \to \uR$, i.e.,
$\cK g = g \circ \bF$,
where $\circ$ represents the composition operator, and $\cK$ is called the Koopman operator \cite{koopman1931hamiltonian} \cite{Budisic12}.
Let $g(t) = g(\bx(t))$ denote the observable at time $t$. Then, it can be shown that
\be
g(\bx(t+1)) = g \circ \bF(\bx(t)) = \cK g(\bx(t)),
\ee
which can also be expressed as $g(t+1) = \cK g(t)$.

As discussed in \cite{Brunton_PLOS}, suppose that any observable  function can be expressed as a weighted sum of $M$ observable functions, $\{g_1, \ldots, g_M\}$, i.e., $g(t) = c_1 g_1 (t) + \ldots + c_M g_M (t)$, where $c_m$ represents the $m$th coefficient. 
Then, a Koopman-invariant subspace can be defined as the spanc of a set of functions, $\{g_1, \ldots, g_M\}$.
Since $\cK$ is a linear operator, it can be shown that $\cK g(t) = 
\sum_m \tilde c_m g_m (t)$ 
for a different set of the coefficients, 
$\{\tilde c_m\}$.
As a result, an $M$-dimensional linear operator, denoted by $\bK$, can be used for the Koopman operator within the Koopman-invariant subspace. The resulting matrix, $\bK$, is called the Koopman matrix \cite{Brunton22}, yielding, 
\be 
\bg (t+1) = \bK \bg(t),
    \label{EQ:linearlonterm}
\ee 
where $\bg (t) = [g_1 (t) \ \ldots \ g_M(t)]^\rT$ is the observable vector.

\subsection{Koopman Operator for Long-Term Prediction}

From \eqref{EQ2:xFx}, the state $\bx (t+\tau)$ of the future time steps $\tau=1, 2, \cdots$ is cast as
\be
\bx (t+\tau) = \bF ( \cdots \bF (\bx(t)) \cdots ) = \bF^\tau (\bx(t)). \label{EQ:longterm}
\ee
For an unknown $\bF$, estimating $\bF$ is crucial for accurate long-term prediction. As reviewed in \cite{Aguirre2009}, existing methods commonly estimate $\bF$ based on past state observations or linearize $\bF$ at a reference point. On the other hand, the Koopman operator offers an approach to prediction in a Koopman invariant subspace. In the Koopman invariant subspace where \eqref{EQ:linearlonterm} holds, \eqref{EQ:longterm} is recast as
$\bg (t+\tau) =  \bK^\tau \bg(t), \ \tau = 1,2,\ldots$.


In \cite{lusch2018deep} \cite{Takeishi17}, an NN is used to derive a mapping for the observable vector, i.e., 
$$
\bg(t) = \varphi (\bx(t)),
$$
where $\varphi$ represents the multi-layer NN. In addition, another NN is considered to recover $\bx(t)$ from $\bg(t)$, resulting in KAE.
In \cite{Azencot20a}, it is demonstrated that the approach presented in \cite{lusch2018deep} does not yield satisfactory long-term predictions. Consequently, an alternative approach, based on the notion of consistent dynamics, is introduced. Notably, this alternative method encompasses both forward and backward predictions, where the backward prediction is carried out with a different linear operator instead of $\bK$ with ensuring consistency. Further elaboration on the approach proposed in \cite{Azencot20a} will be provided in subsequent sections.


Long-term prediction is typically applicable to energy-conserving systems, such as quasi-periodic ones, where time evolution persists. In these cases, the Fourier transform of the output signal sequence may exhibit multiple Dirac delta functions in the frequency domain. Consequently, \cite{Lange21} models the signal as a nonlinear function of sinusoidal inputs (sine and cosine), which may correspond to $\bK$ with eigenvalues on the unit circle.


\section{KAE for Long-Term Prediction}

\subsection{Vanilla KAE}

\begin{figure}[thb]
\begin{center}
\includegraphics[width=0.85\columnwidth]{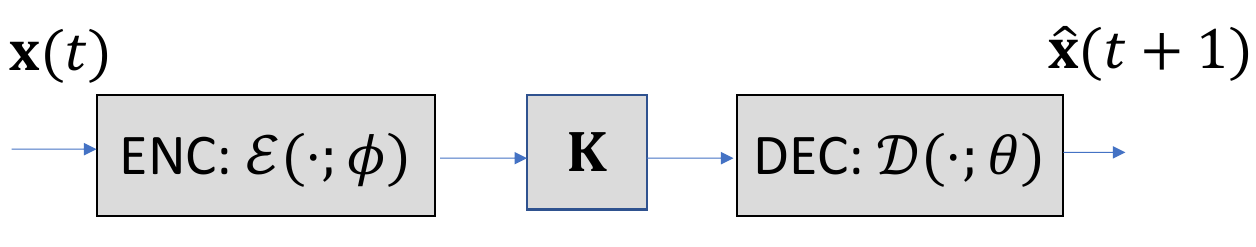} 
\end{center}
\caption{An illustration of KAE.}
        \label{Fig:KAE}
\end{figure}

Suppose that the measurements, denoted by $\bx(t)$, are given and assume that $\bx(t)$ is the state of a nonlinear dynamical system as shown in \eqref{EQ2:xFx}.
Throughout the paper, we consider a KAE that consists of two deep NNs (DNNs) for encoder and decoder, denoted by $\cE$ and $\cD$, respectively, and the Koopman matrix, $\bK$, as illustrated in Fig.~\ref{Fig:KAE}. The input of the KAE is $\bx(t)$ and the output is an estimate of $\bx(t+1)$. 
In particular, the inputs and outputs of the three elements of the KAE and their relationships are given by
\begin{align}
    \label{EQ:K1}
\bz (t) & = \cE(\bx(t); \phi) \in \uR^M \\
    \label{EQ:K2}
\bz (t+1) & = \bK \bz (t)  \in \uR^M \\
    \label{EQ:K3}
\hat \bx (t+1) & = \cD (\bz(t+1); \theta) \in \uR^{N},
\end{align}
where $\bz(t)$ is the output of the encoder with the parameter vector $\phi$, and $\theta$ is the parameter vector of the decoder. Here, $M$ represents the dimension of the latent space and $\bz(t)$ is referred to as the latent vector. In addition, for convenience, $\bx(t)$ will be referred to as the input vector.

The encoder, decoder, and Koopman matrix can be obtained by minimizing the prediction error, which is given by
\be 
F_{T} (\phi, \theta, \bK) = \frac{1}{T} \sum_{t=0}^{T-1} ||\bx (t+1) -\hat \bx(t+1) ||^2  ,
    \label{EQ:opt}
\ee
where $T$ is the number of samples.
Note that it is also necessary to ensure that the pair of the encoder and decoder is designed to reconstruct the input to the encoder. Thus, the reconstructed version of $\bx (t)$,
denoted by $\tilde \bx(t)$, needs to be sufficiently close to $\bx(t)$ as follows:
\be 
\tilde \bx (t) = \cD (\cE(\bx;\phi) ; \theta) \approx \bx(t).
\ee 
Let the reconstruction error be 
\be
R_{T} (\phi, \theta) = \frac{1}{T} \sum_{t=0}^{T-1} ||\bx (t) - \tilde \bx (t)||^2 .
\ee 
Then, the optimization problem to minimize both the reconstruction and prediction 
errors becomes
\be 
\min_{\phi, \theta, \bK} \omega_{\rm id} R_{T} (\phi, \theta)
+ \omega_{\rm f} F_{T} (\phi, \theta, \bK),
    \label{EQ:opt_id}
\ee
where $\omega_{\rm id}$ and $\omega_{\rm f}$ are the weights of the reconstruction and prediction errors, respectively.

As in \cite{Takeishi17} \cite{lusch2018deep} \cite{Brunton22}, the DNN can be used for the encoder and decoder. The Koopman matrix, encoder, and decoder can be obtained by training to minimize the loss function in \eqref{EQ:opt_id}.


\subsection{Vanilla KAE for Long-Term Prediction}

A salient feature of the Koopman operator is its ability to enable long-term predictions through linearization. Consequently,  a $\tau$-step forward prediction,
where $\tau$ is a positive integer, can be performed as follows:
\begin{align}
\hat \bx_\tau (t+\tau) 
& = {\rm Pred}_\tau (\bx(t)) \cr 
& = \cD (\bK^\tau \bz(t); \theta) = \cD (\bK^\tau \cE(\bx(t); \phi); \theta).
    \label{EQ:x_tau}
\end{align}
Thanks to \eqref{EQ:K2}, \eqref{EQ:x_tau} is recast as $\bz(t+\tau)=\bK^\tau \bz(t)$, which is the input to the decoder to obtain an estimate of $\bx (t+\tau)$. 
If the Koopman operator can be reliably represented in a finite basis and a good Koopman embedding can be found by solving \eqref{EQ:opt_id}, the $\tau$-step forward prediction in \eqref{EQ:x_tau} is expected to be reasonably accurate. However, for a large $\tau$ representing long-term prediction, errors are likely to grow. To see this, assume that $||\bx (t)||^2 < c$ for all $t$, where $c$ is a constant, while there exists at least one eigenvalue of $\bK$ with a magnitude greater than 1. Then, as $\tau$ increases, $||\bK^\tau \bz(t)||^2 = ||\bK^\tau \cE(\bx(t); \phi)||^2$ will increase. This implies that $||\hat \bx_\tau (t+\tau) ||^2$ can grow arbitrarily large as $\tau$ becomes large, even though $||\bx (t)||^2 < c$ is assumed.
To avoid this problem, additional terms 
$||\bx (t+\tau) - \hat \bx_\tau (t+\tau)||^2$, where $\tau \ge 2$, can be added to the loss function.

Consequently, the encoder, decoder, and Koopman matrix can be trained to minimize all the prediction errors up to $W$-step  as follows:
\be 
\min_{\phi, \theta, \bK} 
\omega_{\rm id} R_{T} (\phi,\theta) + \omega_{\rm f} F_{T,W} (\phi, \theta, \bK),
    \label{EQ:opt_F}
\ee
where
\be 
F_{T,W} (\phi, \theta, \bK)= \frac{1}{T W} \sum_{t=0}^{T-1} \sum_{\tau=1}^W ||\bff_\tau (t+\tau) ||^2.
\ee
Here, $W \ge 1$ and $\bff_\tau (t+\tau) = \bx(t+\tau) - \hat \bx_\tau (t+\tau)$.
A large $W$ is expected for long-term prediction.

\subsection{Consistent KAE}

Thanks to the linearization, it is also possible to perform the backward prediction. To this end, \eqref{EQ:K2} can be replaced with
\be 
\bz(t - 1) = \bG \bz(t) ,
    \label{EQ:K2_b}
\ee 
where $\bG$ is referred to as the backward Koopman operator.
The backward prediction error can be defined as
\be 
\bb(t) = \bx(t) - \check  \bx (t),
\ee 
where
\be 
\check \bx (t-1) = 
\cD (\bG \cE(\bx(t); \phi); \theta) = (\cD \circ \bG \circ \cE) \bx(t).
\ee 
A generalized backward prediction can also be formulated. That is, since the $\tau$-step backward prediction is given by
\be 
\check \bx_\tau (t-\tau)  = \cD (\bG^\tau \cE(\bx(t); \phi); \theta),
\ee
letting
\be 
B_{T,W} (\phi, \theta, \bG)= \frac{1}{T W} \sum_{t=1}^{T} \sum_{i=1}^W ||\bb_i(t-i) ||^2,
\ee
where $\bb_i(t-i) = \bx(t-i) - \check \bx_i (t-i)$, the encoder, decoder, and backward Koopman matrix can be trained to minimize all the prediction errors up to $W$-step as follows:
\be 
\min_{\phi, \theta, \bG} \omega_{\rm id} R_{T} (\phi, \theta) + 
\omega_{\rm b} B_{T,W} (\phi, \theta, \bG).
\ee

From \eqref{EQ:K2}, if $\bK$ is nonsingular, we have
\be 
\bz (t) = \bK^{-1} \bz(t+1),
\ee 
which means that $\bG$ in \eqref{EQ:K2_b} is $\bK^{-1}$. 
Thus, when combining the forward and backward prediction, we need to the following constraint:
\be 
\bG \bK \approx \bI.
    \label{EQ:GKI}
\ee 
Accordingly, the encoder, decoder, $\bK$, and $\bG$ can be trained to minimize the overall cost function as follows:
\begin{align} 
& \min_{\phi, \theta, \bK, \bG} 
\omega_{\rm id} R_T (\phi, \theta) +
\omega_{\rm f} F_{T,W} (\phi, \theta, \bK) \cr 
& \qquad + 
\omega_{\rm b} B_{T,W} (\phi, \theta, \bG) +
 \omega_{\rm c} C_1 (\bK, \bG), \label{Eq:id2}
\end{align} 
where 
\be
C_1 (\bK, \bG) = ||\bG \bK - \bI ||_{\rm F}^2.
    \label{EQ:Econ}
\ee 
The resulting KAE is referred to as the CKAE due to \eqref{EQ:Econ} \cite{Azencot20a}.

In \eqref{EQ:opt_F}, as mentioned earlier, a large $W$ is required for long-term prediction. However, when $W$ is large, the eigenvalues of $\bK$ should remain within the unit circle, which may limit prediction performance for quasi-periodic $\bx(t)$, as it is desirable for the eigenvalues to be close to the unit circle. Thus, determining the appropriate value for $W$ is challenging--it should neither be too large nor too small.

On the other hand, the CKAE indirectly controls the eigenvalues. As the eigenvalues of $\bK$ approach the origin, those of $\bG$ will have magnitudes greater than 1, leading to unstable backward prediction. The consistency condition in \eqref{EQ:Econ}, i.e., $\bG \bK = \bI$, ensures that the eigenvalues of $\bK$ and $\bG$ lie close to the unit circle when both forward and backward predictions are stable (each corresponding Koopman matrix should have eigenvalues within the unit circle for stable systems). This property makes the CKAE better suited for long-term prediction of quasi-periodic signals, as demonstrated in \cite{Azencot20a}.


\section{KAE via Singular Value Decomposition}

In this section, we propose a novel KAE architecture that effectively controls eigenvalues to enable long-term prediction for quasi-periodic signals. The key idea is to apply SVD to $\bz(t)$ for controlling eigenvalues, as $\bx(t)$ being quasi-periodic implies that $\bz(t)$ is also quasi-periodic, given that the encoder and decoder are memoryless nonlinear functions. This approach is elaborated as follows.

\subsection{Singular Value Decomposition of Koopman Matrix}

As discussed in \cite{Williams15}, long-term dynamics can be captured by the eigenvectors associated with eigenvalues close to the unit circle. Furthermore, when $\bx(t)$ exhibits quasi-periodic behavior, so does $\bz(t)$, and the eigenvalues are expected to be on the unit circle \cite{Lange21}. As a result, for long-term prediction, it is desirable to have the eigenvalues of the Koopman matrix located on the unit circle \cite{Azencot20a}. However, in general, directly controlling the eigenvalues of $\bK$ during training for the KAE can be challenging. Even though $\bx(t)$ is real-valued, the eigenvalues of $\bK$ can be complex-valued, which is often not suited for real-valued standard NN architectures. To detour this issue, in \cite{Azencot20a}, the consistency constraint in \eqref{EQ:Econ} is considered with backward prediction as a means to indirectly force the eigenvalues closer to the unit circle. 

Alternatively, in this subsection, we introduce a novel approach to designing a KAE for long-term prediction via SVD. Prior to discussing the SVD of $\bK$, we can consider a simple approach where the eigenvalues of $\bK$ all lie on the unit circle. Since $\bK$ is a square matrix, if we impose $\bK$ to be a unitary matrix, all the eigenvalues lie on the unit circle. However, in this case, $\bK$ cannot capture any transient behaviors corresponding to small eigenvalues. Nonetheless, we can observe that $\bK$ suitable for long-term prediction might be close to unitary. 


To illustrate, consider the SVD of $\bK$, which is given as
\be
\bK = \bU \bSigma \bV^\rT ,
\ee
where $\bSigma = {\rm diag} (\sigma_1, \ldots, \sigma_M)$, $\bU = [\bu_1 \ \ldots \ \bu_M]$, and $\bV = [\bv_1 \ \ldots \ \bv_M]^\rT$. Here, $\sigma_m$ represents the $m$th singular value of $\bK$, and $\bu_m$ and $\bv_m$ are the left and right singular vectors of $\sigma_m$, respectively. For long-term prediction, it is desirable to have the eigenvalues of $\bK$ close to the unit circle as discussed earlier (or $\bK$ to be close to a unitary matrix). Thus, if $\bK$ has only eigenvalues on the unit circle, it can be shown that
\be 
\sigma_m = 1, \ m = 1,\ldots, M,
    \label{EQ:sl}
\ee 
yielding another important loss term for long-term prediction:
\begin{align}
V(\bSigma) = ||\bSigma - \bI||^2.
\end{align}
In addition, to ensure that $\bU$ and $\bV$ are unitary, the following loss term has to be included:
\begin{align}
S(\bU, \bV) = ||\bU^\rT \bU - \bI||^2 +   ||\bV^\rT \bV - \bI||^2.
\end{align}




{\color{red} 

}

\subsection{Consistency SVD-based KAE }

\begin{figure}[thb]
\begin{center}
\includegraphics[width=0.85\columnwidth]{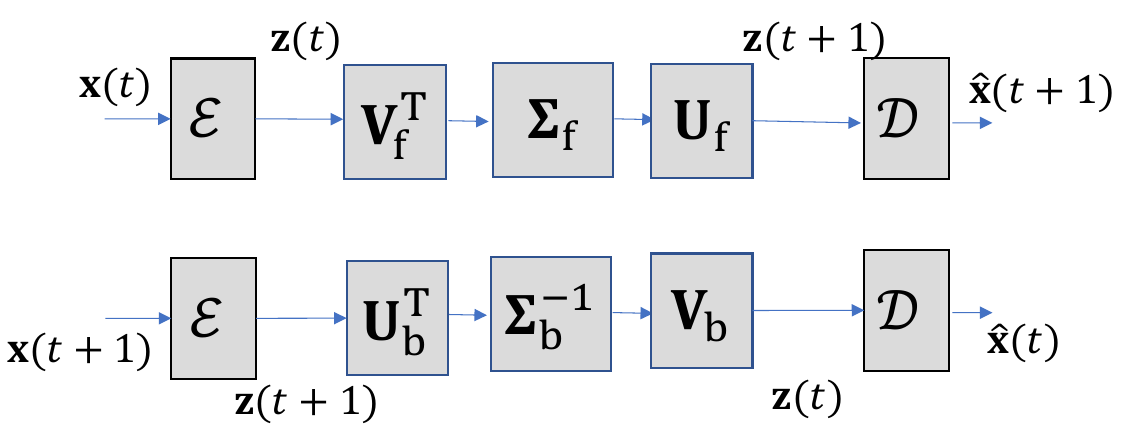} 
\end{center}
\caption{A pair of KAEs with forward and backward prediction for the consistent SVD-based KAE.}
        \label{Fig:cross}
\end{figure}

Using the SVD, it is possible to force $\bK$ to be close to a unitary matrix (that has all eigenvalues on the unit circle). This approach can be extended with the backward prediction.
To this end, for convenience, let $\bSigma_{\rm f} = \bSigma$,
$\bU_{\rm f} = \bU$, and $\bV_{\rm f} = \bV$.
For the backward prediction, we can have the following backward Koopman matrix:
\be 
\bG = \bV_{\rm b} \bSigma^{-1}_{\rm b} \bU^\rT_{\rm b}.
    \label{EQ:GKi}
\ee 
Since $\bG = \bK^{-1}$, we expect to have $\bU_{\rm b} = \bU_{\rm f}$,
$\bV_{\rm b} = \bV_{\rm f}$, and $\bSigma_{\rm b} = \bSigma_{\rm f}$.
The resulting consistent SVD-based KAE can be illustrated as in Fig.~\ref{Fig:cross}.

Finally, the aggregate loss function can be given by
\begin{align}
& \cL = \omega_{\rm id} R_T (\phi, \theta) + \omega_{\rm f} F_{T,W} (\phi, \theta, \bK)  \cr
& + \omega_{\rm b} B_{T,W} (\phi, \theta, \bG)
+ \omega_{\rm sv} V(\bSigma_{\rm f}, \bSigma_{\rm b}) \cr
& + \omega_{\rm c} \left( S (\bU_{\rm f}, \bV_{\rm f}) +S  (\bU_{\rm b}, \bV_{\rm b}) + C (\bV_{\rm f},\bV_{\rm b},\bU_{\rm f},\bU_{\rm b}) \right) ,\qquad \quad
    \label{EQ:loss}
\end{align}
where
\begin{align}
V (\bSigma_{\rm f}, \bSigma_{\rm b} ) & = ||\bI -\bSigma_{\rm f} ||^2 +  
 ||\bI -\bSigma_{\rm b} ||^2 \cr 
C (\bV_{\rm f},\bV_{\rm b},\bU_{\rm f},\bU_{\rm b}) & = ||\bV_{\rm f} \bV_{\rm b}^\rT - \bI||^2
+ ||\bU_{\rm f} \bU_{\rm b}^\rT - \bI||^2
. \quad \quad
\end{align}
In \eqref{EQ:loss}, the first three terms on the right-hand side (RHS) are identical to those for the CKAE \cite{Azencot20a}. The next terms are due to the SVD of $\bK$ and $\bG$.
In particular, the last term is to ensure that $\bU_{\rm f} \approx 
\bU_{\rm b}$ and $\bV_{\rm f} \approx \bV_{\rm b}$, while they are unitary.

\section{Experiments}   \label{S:EXP}
In this section, we provide the results of experiments for evaluating the performance of our proposed approach. Furthermore, we compare the proposed method with available state-of-the-art baseline methods. 
In USVD-KAE, we extend the CKAE architecture by decomposing the Koopman matrix $\mathbf{K}$ into three components: $\mathbf{U}_\text{f}$ and $\mathbf{V}^\rT_\text{f}$ are represented using a trainable matrix via a linear layer, while $\Sigma_\text{f}$ is represented using a trainable diagonal matrix. Similarly, we employ a similar architectural design for the backward prediction matrix decomposition, $\mathbf{G} = \mathbf{V}_\text{b} \Sigma_{\text{b}} \mathbf{U}^\rT_\text{b}$, where $\mathbf{G} = \mathbf{K}^{-1}$. This framework enables direct control over the SVD components, leveraging the unitary properties of the matrices to enhance long-term prediction.

\textbf{Simulation Settings}.\quad 
We compare our proposed USVD-CKAE to the following baselines: 1) Vanilla KAE \cite{lusch2018deep} with the loss function in \eqref{EQ:opt_id}; and 2) CKAE \cite{Azencot20a} with the loss function in \eqref{Eq:id2}. Furthermore, to demonstrate the impact of SVD on performance, we additionally consider a baseline termed 3) iterative SVD-CKAE (ISVD-CKAE), which performs SVD after each training round. Note that ISVD-CKAE has the same loss function \eqref{EQ:loss} as USVD-CKAE, but may incur a higher computing cost due to iterative SVD computations unlike USVD-CKAE. 
To enable comparison, we define our performance metric, where the prediction error at the $\tau$-th time step is given by:
\be
\epsilon_\text{pred}(\tau) = ||\mathbf{x}(t_0+p) - \mathcal{D}(\mathbf{K}^\tau \mathbf{z}(t_0))||^2,
\ee
where $t_0$ is the index of the initial step. Then, the average prediction error is given as $\bar{\epsilon}_\text{pred}(P) = \frac{1}{P} \sum_{\tau = 1}^P \epsilon_\text{pred}(\tau)$, where $P$ is the index of the final prediction step. For simulations, we consider fluid dynamics explores the behavior of fluids, governed by the complex Navier-Stokes equations, which are nonlinear partial differential equations. In line with the approach outlined in \cite{Ipiss}, a practical application of the Galerkin method is employed to simulate fluid flow around a circular cylinder. This method simplifies the problem by introducing a Reynolds number (Re = 100) for creating 3-dimensional dynamics over 1500 uniformly spaced points, such that $\mathbf{x}(t)\in \mathbb{R}^3$.

\begin{table}
\centering
\caption{Simulation parameters for the proposed architecture.}
\resizebox{.9\columnwidth}{!}{\begin{tabular}{ll}
\toprule
\textbf{Parameter} & \textbf{Value} \\
\midrule
Loss weights ($\{\omega_{\text{id}}, \omega_{\text{f}}, \omega_{\text{b}}, \omega_{\text{sv}}, \omega_{\text{c}}\}$) & \{$1, 1, 10^{-2}, 10^{-4}, 1$\} \\
Forward Prediction Steps ($W_\text{f}$) & $20$ \\
Backward Prediction Steps ($W_\text{b}$) & $20$ \\
Batch size & $16$ \\
Learning rate & $10^{-2}$ \\
\bottomrule
\label{tab:SP2}
\end{tabular}}
\end{table}

\begin{figure}
    \centering
    \includegraphics[width=0.5\textwidth]{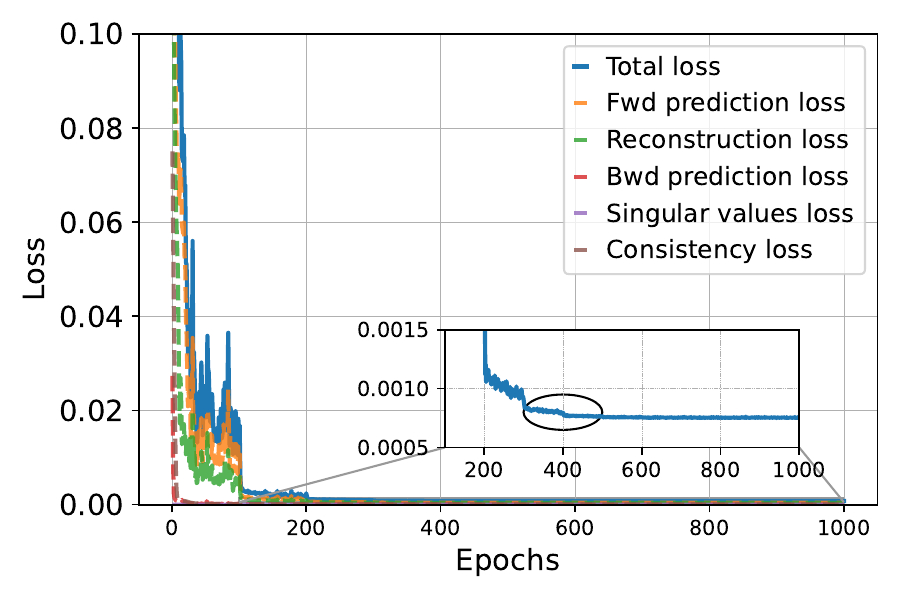}
    \caption{Loss curve evolution during the training of our proposed method USVD-CKAE.}
    \label{fig:m1}
\end{figure}

\textbf{Training Convergence}.\quad
In Fig.~\ref{fig:m1}, we illustrate the training loss curve of our proposed method, USVD-CKAE, showcasing both the total weighted loss and its individual components. Notably, USVD-CKAE converges around the $400$th epoch, which is $2.5$ times faster than ISVD-CKAE, despite having the same loss function. It is presumably because USVD-CKAE directly updates eigenvalues in diagonal matrices $\mathbf{\Sigma}_{\rm f}$ and $\mathbf{\Sigma}_{\rm b}$, whereas ISVD-CKAE adjusts these values only through $\mathbf{K}$. Consequently, USVD-CKAE achieves a lower computing cost than ISVD-CAKE, thanks to its fast convergence and avoiding SVD computations, as we shall elaborate next.




\begin{table}
\centering
\caption{Prediction errors ($\bar{\epsilon}_\text{pred}(1, 1001)$) and compute~cost.}
\label{tab:approach_comparison}
\resizebox{.8\columnwidth}{!}{\begin{tabular}{l c c}
\toprule
\textbf{Method} & \textbf{Pred. Error} & \textbf{Comput. Cost} (FLOPS) \\
\midrule
USVD-CKAE &  0.00047  & 168 M\\
\midrule
ISVD-CKAE & \textbf{0.00035} & 362 M\\
\midrule
Vanilla KAE        & 0.0014 & \textbf{43 M}\\
\midrule
CKAE       & 0.00523 & 100 M\\
\bottomrule
\end{tabular}}
\label{table:k}
\end{table}

\begin{figure}
    \centering
    \vspace{10pt}
    \includegraphics[trim={0 5.5cm 0 0.7cm}, clip, width=.9\columnwidth]{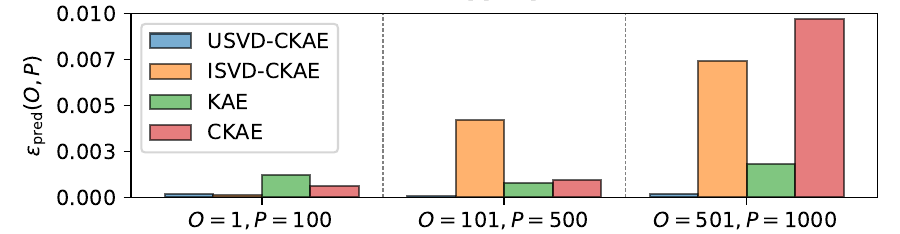}
    \caption{Prediction error comparison at $200$ epochs.}
    \label{fig:PredEarly}
\end{figure}

\textbf{Average Prediction Errors and Computing Costs}.\quad
We provide the average prediction error over 1000 steps and its computational complexity measured in terms of floating-point operations per second (FLOPS) in Table~\ref{table:k}. USVD-CKAE surpasses the performance of Vanilla KAE and CKAE by 66.43\% and 91.01\%, respectively. Moreover, USVD-CKAE achieves a comparable prediction performance to ISVD-CKAE while demanding 53.6\% fewer computational operations in terms of FLOPS. Given limited computation, thanks to its faster convergence, USVD-CKAE achieves even a lower prediction error than ISVD-CKAE, as evidenced by Fig.~\ref{fig:PredEarly} where training is limited by $200$ epochs.

\begin{figure}
    \centering
        \begin{subfigure}[b]{0.23\textwidth}
        \centering
        \includegraphics[width=\linewidth]{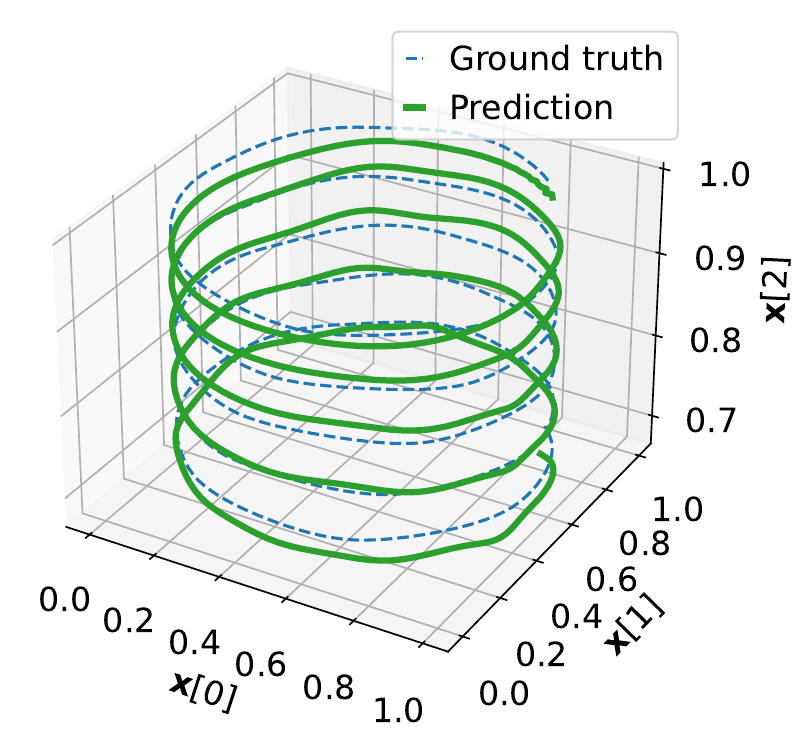}
        \caption{Comparison of predicted and the ground truth for 1000-steps.}
        \label{Fig:pred2}
    \end{subfigure}
    \hfill
    \begin{subfigure}[b]{0.24\textwidth}
        \centering
        \includegraphics[width=\linewidth]{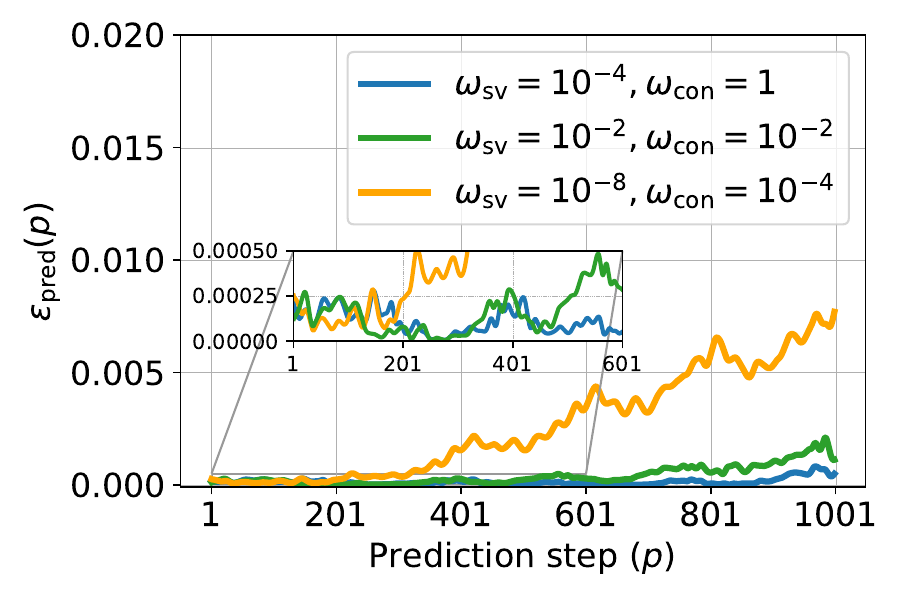}
        \caption{Evolution of prediction error over 1000-step predictions using different settings of loss weights.}
        \label{Fig:pred1}
    \end{subfigure}
    \vspace{10pt}
    \begin{subfigure}[b]{0.48\textwidth}
        \centering
        \includegraphics[width=\linewidth]{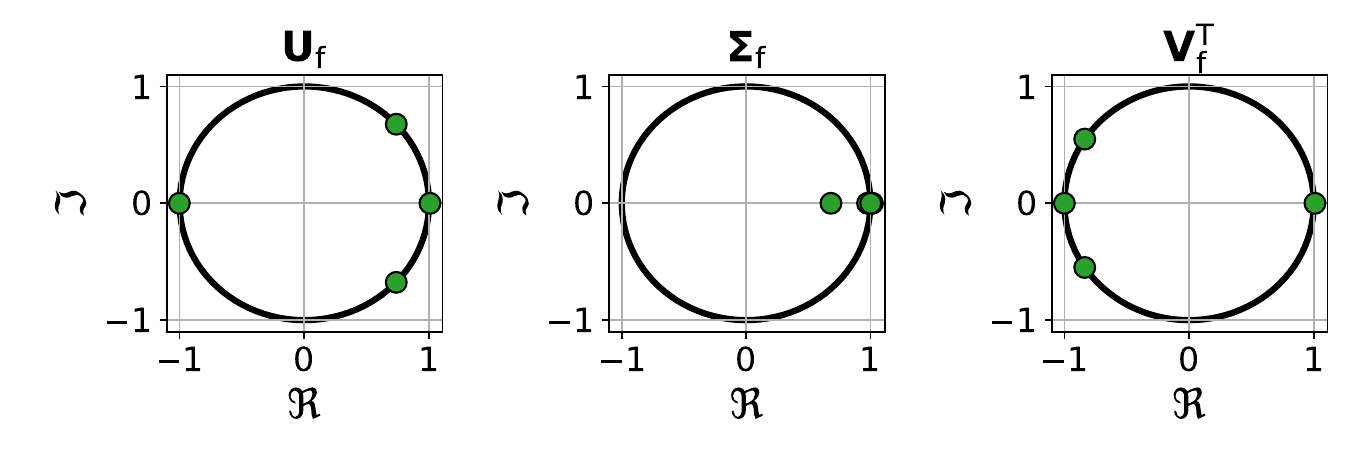}
        \caption{Visualizing the eigenvalues of the unrolled SVD components for the forward prediction matrix $\mathbf{K}$.}
        \label{Fig:pred3}
    \end{subfigure}
    \caption{Evaluating the predictive performance of our proposed method using performance metrics and trajectory comparison with the ground truth.} 
    \label{Fig:e}
\end{figure}

\textbf{Impact of Eigenvalue Control}.\quad
Fig.~\ref{Fig:pred2} visualizes the error of future $1000$-step prediction compared to the ground truth dynamics. To enable such a long-term prediction, it is necessary for $\mathbf{K}$ to demonstrate unitary properties, maintaining norm-preserving iterative multiplication over a vector. The influence of fostering unitary matrices is controlled by the weights $\omega_{\text{c}}$ and $\omega_{\text{sv}}$ in \eqref{EQ:loss}. Fig.~\ref{Fig:pred1}, we assess the prediction error over 1000-steps across three settings, each varying based on the loss weights. 
A greater emphasis on enforcing unitary properties leads to a decrease in long-term prediction errors. In contrast, lower control settings typically result in an increase in predictive errors over time, albeit with improved short-term accuracy.
In Fig.~\ref{Fig:pred3}, we analyze the unitary properties of the forward prediction components by visualizing their eigenvalues. Unitary matrices have eigenvalues lying on the unit circle, denoted as $|\lambda_i| = 1$ for $i = 1, \cdots, M$, which is shown for all matrices. While the eigenvalues of $\Sigma_{\text{f}}$ may not strictly adhere to the unit circle, increasing the weight $\omega_{\text{sv}}$ can enforce this. However, such strict control may degrade performance of the short-term predictions.


\section{Conclusions}

For long-term prediction, in this paper, we explored the application of the SVD of the Koopman matrix and the inclusion of backward prediction for consistency to develop a data-driven approach. By introducing a loss term for singular values in the resulting Koopman autoencoder, we were able to manipulate these values, ensuring that the eigenvalues of the Koopman matrix remain close to the unit circle--a critical factor for effective long-term prediction. Experimental results validated the effectiveness of this approach, demonstrating that the proposed Koopman autoencoder outperforms baseline methods in long-term prediction tasks.

\vfill
\pagebreak

\bibliographystyle{IEEEbib}
\bibliography{Koopman}

\begin{thebibliography}{10}

\bibitem{koopman1931hamiltonian}
Bernard~O Koopman,
\newblock ``Hamiltonian systems and transformation in hilbert space,''
\newblock {\em Proceedings of the National Academy of Sciences}, vol. 17, no.
  5, pp. 315--318, 1931.

\bibitem{Brunton_PLOS}
Steven~L. Brunton, Bingni~W. Brunton, Joshua~L. Proctor, and J.~Nathan Kutz,
\newblock ``Koopman invariant subspaces and finite linear representations of
  nonlinear dynamical systems for control,''
\newblock {\em PLOS ONE}, vol. 11, no. 2, pp. 1--19, 02 2016.

\bibitem{Brunton22}
Steven~L. Brunton, Marko Budi\v{s}i\'{c}, Eurika Kaiser, and J.~Nathan Kutz,
\newblock ``Modern {K}oopman theory for dynamical systems,''
\newblock {\em SIAM Review}, vol. 64, no. 2, pp. 229--340, 2022.

\bibitem{lusch2018deep}
Bethany Lusch, J~Nathan Kutz, and Steven~L Brunton,
\newblock ``Deep learning for universal linear embeddings of nonlinear
  dynamics,''
\newblock {\em Nature communications}, vol. 9, no. 1, pp. 4950, 2018.

\bibitem{Takeishi17}
Naoya Takeishi, Yoshinobu Kawahara, and Takehisa Yairi,
\newblock ``Learning {K}oopman invariant subspaces for dynamic mode
  decomposition,''
\newblock in {\em Proceedings of the 31st International Conference on Neural
  Information Processing Systems}, Red Hook, NY, USA, 2017, NIPS'17, p.
  1130–1140, Curran Associates Inc.

\bibitem{Azencot20a}
Omri Azencot, N.~Benjamin Erichson, Vanessa Lin, and Michael Mahoney,
\newblock ``Forecasting sequential data using consistent {K}oopman
  autoencoders,''
\newblock in {\em Proceedings of the 37th International Conference on Machine
  Learning}, Hal~Daumé III and Aarti Singh, Eds. 13--18 Jul 2020, vol. 119 of
  {\em Proceedings of Machine Learning Research}, pp. 475--485, PMLR.

\bibitem{Budisic12}
Marko Budišić, Ryan Mohr, and Igor Mezić,
\newblock ``{Applied Koopmanism},''
\newblock {\em Chaos: An Interdisciplinary Journal of Nonlinear Science}, vol.
  22, no. 4, pp. 047510, 12 2012.

\bibitem{Aguirre2009}
Luis~A. Aguirre and Christophe Letellier,
\newblock ``Modeling nonlinear dynamics and chaos: A review,''
\newblock {\em Mathematical Problems in Engineering}, vol. 2009, pp. 1--35,
  2009.

\bibitem{Lange21}
Henning Lange, Steven~L. Brunton, and J.~Nathan Kutz,
\newblock ``From {F}ourier to {K}oopman: Spectral methods for long-term time
  series prediction,''
\newblock {\em J. Mach. Learn. Res.}, vol. 22, no. 1, Jan. 2021.

\bibitem{Williams15}
Matthew~O. Williams, Ioannis~G. Kevrekidis, and Clarence~W. Rowley,
\newblock ``A data{\textendash}driven approximation of the {K}oopman operator:
  Extending dynamic mode decomposition,''
\newblock {\em Journal of Nonlinear Science}, vol. 25, no. 6, pp. 1307--1346,
  June 2015.

\bibitem{Ipiss}
Opal Issan,
\newblock ``Enhancing dynamic mode decomposition using autoencoder networks.,''
  \url{https://github.com/opaliss/dmd_autoencoder}, 2021.

\end{thebibliography}

\end{document}